\documentclass[letterpaper, 10 pt, conference]{ieeeconf}  % Comment this line out if you need a4paper
\usepackage{graphicx}
\usepackage{booktabs}
\usepackage{dblfloatfix} % Улучшает позиционирование figure* и разрешает [b] (bottom)

\IEEEoverridecommandlockouts                              % This command is only needed if 
\usepackage{amsmath} % assumes amsmath package installed

\title{\LARGE \bf
ZeroTouch: Tactile-Supervised Visual Contact \\ Estimation for Contact-Rich Manipulation
}

\author{Dmitriy Kosenkov, Daniia Zinniatullina, Miguel Altamirano Cabrera, \\
Iana Zhura, Mikhail Derevianchenko, and Dzmitry Tsetserukou% <-this % stops a space
\thanks{The authors are with the Intelligent Space Robotics Laboratory, Skolkovo Institute of Science and Technology, Moscow, Russian Federation.
       {\tt\small {Dmitriy.Kosenkov, Daniia.Zinniatullina, M.Altamirano, Iana.Zhura, Mikhail.Derevianchenko, D.Tsetserukou}@skoltech.ru}}%
}
\begin{document}

\bstctlcite{IEEEexample:BSTcontrol}

\maketitle
\thispagestyle{empty}
\pagestyle{empty}

%%%%%%%%%%%%%%%%%%%%%%%%%%%%%%%%%%%%%%%%%%%%%%%%%%%%%%%%%%%%%%%%%%%%%%%%%%%%%%%%
\begin{abstract}
Reliable robotic grasping benefits from estimating the evolving physical interaction and selecting a grasp-dependent compression target. Tactile sensors provide direct interaction measurements but require dedicated hardware at deployment. We introduce ZeroTouch, a tactile-supervised framework that predicts dense contact deformation, the instantaneous six-axis wrench, and a grasp-dependent desired compression target from wrist RGB observations, gripper state, and local gravity direction. Tactile measurements are used only as privileged supervision during training and are not required at deployment. On the full validation set, the complete architecture reduces normal-force MAE from \textbf{2.017}~N for a state-only baseline to \textbf{0.531}~N. In physical evaluation with 20 trials per condition, ZeroTouch achieves \textbf{95\%} success on an unseen object, \textbf{80\%} in a seen-object/unseen-grasp condition, and \textbf{90\%} under a content/load shift. Under the same evaluation protocol, OpenVLA achieves \textbf{25\%}, \textbf{40\%}, and \textbf{55\%}, while SmolVLA achieves \textbf{10\%}, \textbf{25\%}, and \textbf{35\%}, respectively.
\end{abstract}

%%%%%%%%%%%%%%%%%%%%%%%%%%%%%%%%%%%%%%%%%%%%%%%%%%%%%%%%%%%%%%%%%%%%%%%%%%%%%%%%
\section{INTRODUCTION}

Reliable robotic grasping requires not only selecting where to grasp, but also regulating the physical interaction after contact. Insufficient force can cause slip, whereas excessive force can deform or damage fragile objects. The resulting force--torque response also depends on grasp location, gripper orientation, and contact geometry, even for the same object \cite{bicchi2000robotic,calandra2017feeling,bekiroglu2011assessing}. Tactile sensing provides direct contact information, but tactile-equipped grippers require dedicated hardware at deployment \cite{lambeta2020digit}, limiting the applicability of learned policies.

Recent vision--touch research has shown that visual observations can predict tactile and physical object properties. Prior work has estimated haptic properties from vision \cite{gao2016deep,takahashi2019deep}, while large paired datasets such as Touch and Go \cite{yang2022touch} have enabled visuo--tactile representation learning and tactile-signal prediction. However, most existing methods focus on tactile-property estimation, cross-modal generation, or tactile representations rather than the physical interaction state required for force-aware grasping. In particular, jointly estimating contact deformation, the instantaneous 6D force--torque wrench, and a grasp-dependent force target from vision remains underexplored.

This inference is challenging because object appearance alone does not uniquely determine contact mechanics. Prior work has estimated interaction forces from camera observations and robot signals \cite{lee2018interaction}, modeled vision-based grip forces \cite{ko2023vision}, or exploited visual deformation of compliant hands \cite{zhu2025forces}. Nevertheless, similar RGB observations can correspond to different wrench responses depending on contact location and gripper configuration, while the interaction evolves from free space to contact and compression. Models that rely mainly on object appearance may therefore learn object-specific correlations rather than transferable visual cues of physical interaction.

\begin{figure*}[t]
    \centering
    \includegraphics[width=\textwidth]{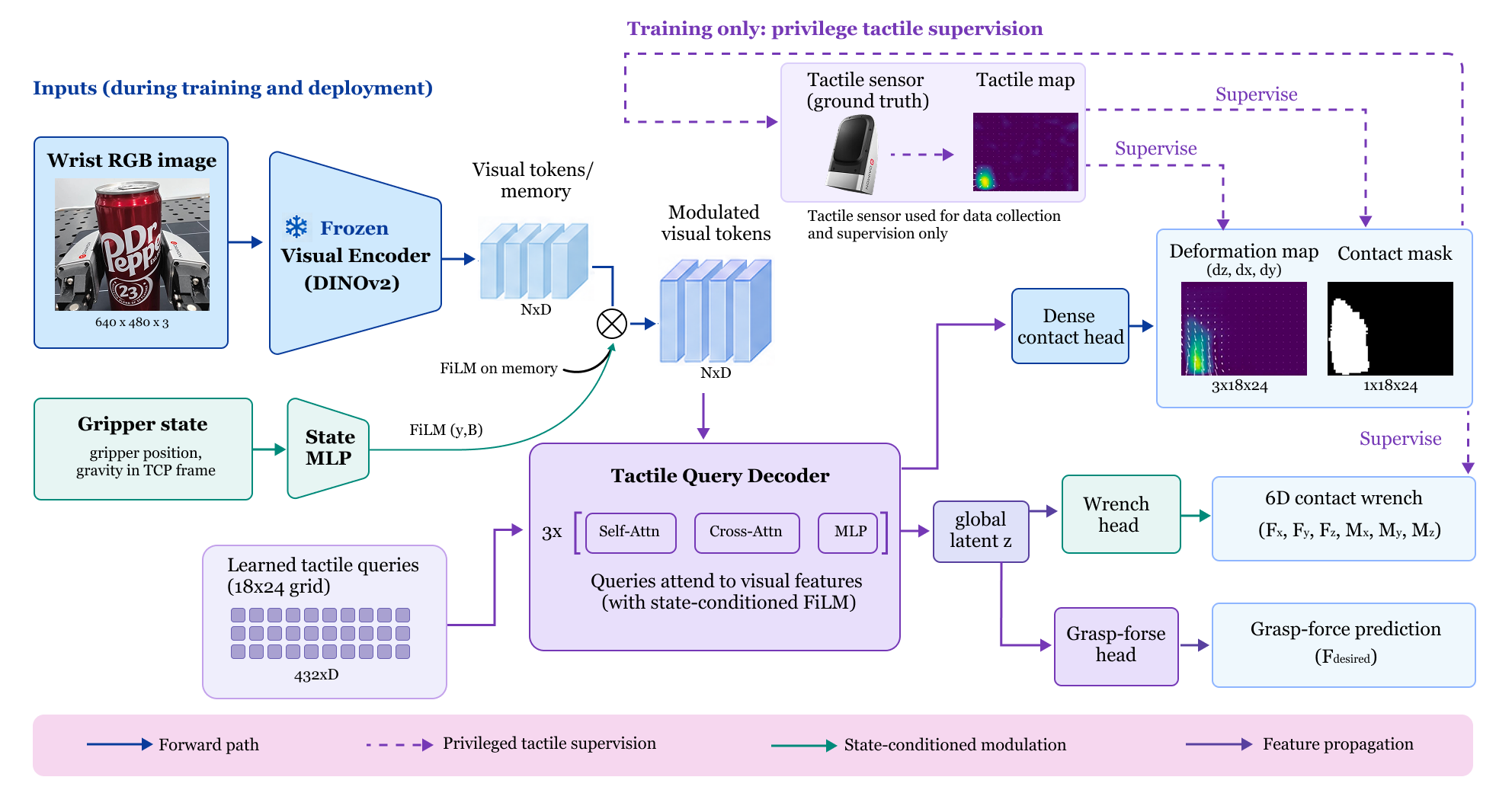}
    \caption{ZeroTouch framework. Wrist RGB observations, gripper state, and local gravity direction are used to predict contact deformation, the instantaneous 6D wrench, and the desired grasp force. Tactile sensing is used only during training as privileged supervision.}
    \vspace{-4mm}
    \label{fig:framework}
\end{figure*}

We introduce \textbf{ZeroTouch}, a tactile-supervised visual framework (Fig.~\ref{fig:overview}) that predicts the physical state of a grasp from wrist-camera RGB observations, gripper state, and the gravity direction expressed in the local gripper frame. The overall architecture is illustrated in Fig.~\ref{fig:framework}.  In addition to the measured gripper position, we provide the normalized gravity vector, which encodes the orientation of the gripper relative to gravity. This proprioceptive signal disambiguates grasp configurations that may appear visually similar but produce different force and torque responses due to orientation. From these inputs, ZeroTouch jointly estimates a dense contact deformation map, the instantaneous 6D wrench, and the desired grasp force for the current object and grasp configuration. Tactile sensing is used only during training as privileged physical supervision; at deployment, the model operates without tactile input.

\begin{figure}[t]
    \centering
    \includegraphics[width=\columnwidth]{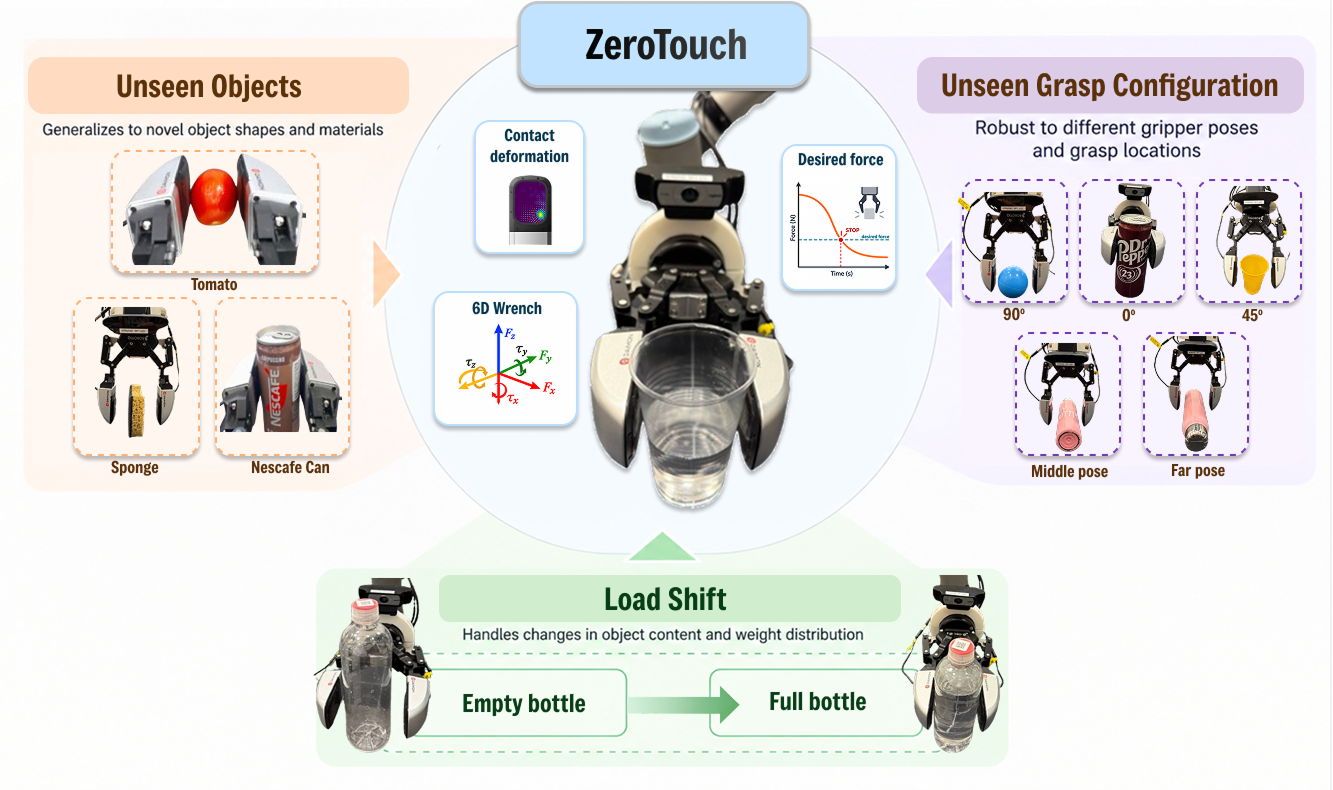}    \label{fig:overview}
    \vspace{-4mm}
    \caption{Overview of ZeroTouch. From wrist RGB, gripper state, and local gravity direction, the model predicts contact deformation, the instantaneous 6D wrench, and the desired grasp force. The framework operates across different gripper poses and grasp locations on the object, while tactile sensing is used only during training.}
    \vspace{-4mm}
    \label{fig:overview}
\end{figure}

ZeroTouch is trained over the full grasp sequence, including non-contact, contact onset, and object compression. This allows the model to associate changes in visual contact geometry, gripper configuration, and gravity-relative orientation with the corresponding evolution of forces and moments rather than relying only on object identity. We evaluate the method across objects with different geometries and physical properties, multiple grasp locations and gripper orientations, and held-out objects and grasp configurations. The main contributions of this work are:

\begin{enumerate}

    \item \textbf{Tactile-supervised tactile-free contact estimation framework} that predicts dense contact deformation and the instantaneous 6D contact wrench from wrist RGB observations, gripper state, and local gravity direction, while requiring no tactile sensing at deployment.

    \item \textbf{Joint grasp-force prediction formulation} that complements the estimated current contact state with a desired grasp force, enabling the model to represent both how the object is currently being loaded and how strongly it should be held.
    
    \item \textbf{Real-world evaluation under object, grasp, and load shifts}, comparing ZeroTouch with vision-language-action baselines in physical grasping experiments with controlled arm motion and differing gripper-closure strategies.
    
\end{enumerate}

\section{RELATED WORK}

ZeroTouch relates to visual--tactile cross-modal learning, vision-based contact estimation, and force-aware manipulation. We build on these directions by using tactile sensing as privileged supervision to jointly predict dense contact deformation, the instantaneous 6D interaction wrench, and a grasp-dependent force target from RGB observations and robot state, without tactile input at deployment.

%\subsection{Visual--Tactile Cross-Modal Prediction}
\subsection{Visual--Tactile Cross-Modal Learning}
Cross-modal methods have shown that visual and tactile information can be transferred between sensing modalities. Li et al.~\cite{li2019connecting} demonstrated bidirectional prediction between vision and GelSight, while Lee et al.~\cite{lee2020making} learned visuo-tactile representations for contact-rich manipulation. More recent approaches exploit tactile information without requiring touch at deployment: VITaL~\cite{george2025vital} uses visuo-tactile pretraining for tactile and vision-only policies, TacImag~\cite{zhang2026tacimag} generates imagined tactile observations from vision and proprioception, and FELT~\cite{li2026felt} predicts tactile pressure representations directly from RGB. In contrast, ZeroTouch couples dense tactile prediction with explicit estimation of the current 6D wrench and a grasp-dependent force target.

\subsection{Visual Estimation of Contact Forces}

Physical interaction can also be inferred directly from visual observations. VPEC~\cite{grady2022visual} estimates contact pressure from RGB and uses it for closed-loop manipulation, while VFTS~\cite{collins2023forcetorque} predicts six-axis force/torque measurements from a single image. ViPER~\cite{collins2023viper} extends visual pressure estimation to diverse operating conditions, and ForceSight~\cite{collins2024forcesight} predicts visual-force goals for manipulation. ZeroTouch differs by jointly learning local tactile deformation, the global interaction wrench, and the grasp-dependent target used to terminate gripper closure within a common tactile-supervised representation.

\subsection{Force-Aware and Tactile-Conditioned Manipulation}
Force and tactile signals have also been incorporated directly into learned manipulation policies. TacDiffusion~\cite{wu2025tacdiffusion} generates 6D wrenches for tactile manipulation, FARM~\cite{helmut2025farm} conditions a diffusion policy on tactile observations to predict motion and gripping force, and FeelTheForce~\cite{adeniji2025feel} predicts contact-force targets for closed-loop control. RETAF~\cite{kang2026force} similarly performs high-frequency force regulation using wrist vision and tactile feedback. Unlike methods that retain tactile or force sensing during execution, ZeroTouch uses these signals only as training supervision and performs grasp-force estimation and termination from vision and robot state alone.

\section{ZeroTouch FRAMEWORK}

\subsection{Problem Formulation}

ZeroTouch addresses force-aware robotic grasping without requiring tactile sensing at deployment. During training, each interaction is represented by the multimodal observation:
\begin{equation}
    \mathbf{o}_t =
    \left\{
    \mathbf{o}^{\mathrm{vis}}_t,
    \mathbf{o}^{\mathrm{tac}}_t,
    \mathbf{o}^{\mathrm{prop}}_t
    \right\},
\end{equation}
where $\mathbf{o}^{\mathrm{vis}}_t$ denotes the wrist-mounted RGB observation, $\mathbf{o}^{\mathrm{tac}}_t$ the tactile observation, and $\mathbf{o}^{\mathrm{prop}}_t$ the gripper state. The proprioceptive observation further includes the normalized gravity direction expressed in the local gripper/TCP frame:
\begin{equation}
    \mathbf{g}_{\mathrm{local},t}
    =
    \left[
    g_x,\,
    g_y,\,
    g_z
    \right],
    \qquad
    \|\mathbf{g}_{\mathrm{local},t}\|_2 \approx 1.
\end{equation}This vector specifies the orientation of the gripper relative to gravity
and helps disambiguate grasp configurations that may appear visually similar but induce different force and torque responses. Tactile observations are available only during training and provide privileged physical supervision. At deployment, tactile measurements are unavailable, and ZeroTouch operates only from the wrist RGB observation and robot state. From these inputs, the model predicts a dense tactile deformation representation $\hat{\mathbf{D}}_t$, the instantaneous six-axis interaction wrench:
\begin{equation}
\hat{\mathbf{w}}_t =
\left[
\hat F_x,
\hat F_y,
\hat F_z,
\hat M_x,
\hat M_y,
\hat M_z
\right],
\end{equation}
and a signed desired normal-force target $\hat{F}^\star$.

The two force quantities have different roles. The instantaneous estimate $\hat F_z(t)$ describes the current normal interaction force and becomes more negative as compression increases. In contrast, $F^\star$ is an episode-level desired-force target associated with the selected grasp configuration and is estimated from pre-contact visual observations.

During data collection, the gripper closure was adjusted for each grasp configuration until the object could be lifted by $5,\mathrm{cm}$, held for $2,\mathrm{s}$ without observable slip, and subsequently returned. Only grasp trials satisfying this lift-and-hold criterion were retained as successful demonstrations for desired-force supervision.

We observed that the normal-force signal can exhibit a transient peak immediately after gripper closure, followed by a lower approximately steady value even though the measured gripper position remains unchanged. We therefore do not define the desired force from the instantaneous peak. Instead, for each retained grasp trajectory, we compute:
\begin{equation}
F^\star =
\operatorname{median}_{t\in\mathcal{T}_{\mathrm{settle}}}
F_z(t),
\end{equation}
where compression corresponds to negative sensor $F_z$, and $\mathcal{T}_{\mathrm{settle}}$ denotes the post-closure interval in which the gripper position is stable and valid contact is maintained. Thus, stronger compression corresponds to a more negative value of $F^\star$.

We refer to $F^{\star}$ as the \emph{desired grasp-compression force}.  Operationally, it represents the settled compression measured for a grasp that satisfied the standardized lift-and-hold criterion during data collection. This definition is empirical: it does not claim a globally optimal force or a theoretically guaranteed minimum-safe force. Rather, it provides a physically interpretable grasp-level target grounded in successful object retention under the evaluated protocol.

The resulting deployment-time prediction is:
\begin{equation}
\left(
\mathbf{o}^{\mathrm{vis}}_{t},
\mathbf{o}^{\mathrm{prop}}_{t},
\mathbf{g}_{\mathrm{local},t}
\right)
\rightarrow
\left(
\hat{\mathbf{D}}_t,
\hat{\mathbf{w}}_t,
\hat F^{\star}
\right).
\end{equation}

ZeroTouch focuses on contact-state estimation and grasp-force regulation. The approach motion and grasp pose are supplied by an upstream manipulation policy or scripted controller and are outside the scope of the proposed model.

\subsection{Tactile-Supervised Contact Representation}

Given the wrist-camera observation $\mathbf{o}^{\mathrm{vis}}_t$ and robot state $\mathbf{o}^{\mathrm{prop}}_t$, ZeroTouch first extracts a visual representation of the current gripper--object configuration. The RGB observation is encoded using a frozen DINOv2 visual backbone, producing patch-level visual tokens. These tokens are linearly projected to the decoder dimension.

In parallel, the measured gripper position and the normalized gravity direction in the local TCP frame are encoded by a lightweight state MLP. The resulting state embedding produces feature-wise scaling and bias parameters that modulate the visual tokens through FiLM before cross-attention. Conditioning the visual memory before attention allows the gripper configuration to affect not only the magnitude of the output but also the spatial visual features used for contact inference.

To recover spatial contact information, we introduce a set of learnable tactile queries arranged on an $18\times24$ grid. The queries are processed by repeated self-attention, cross-attention, and MLP blocks. Self-attention allows neighboring tactile queries to exchange information, while cross-attention associates each tactile location with relevant visual patches.

The resulting spatial query features are reshaped into a dense feature grid and decoded into tactile deformation components $(d_z,d_x,d_y)$ and a contact representation. In parallel, a dedicated global interaction token is used to form a latent representation $\mathbf{z}_t$ for global mechanical quantities. This representation is provided to the wrench and desired-force prediction heads.

During training, the tactile sensor provides privileged supervision for the dense deformation representation and the recorded six-axis wrench. The desired-force branch is supervised only on eligible pre-contact frames using the grasp-level target $F^{\star}$ defined above. At deployment, no tactile observation or recorded force measurement is supplied to the model.
\subsection{Vision-Based Grasp Termination}

ZeroTouch controls only gripper closure; the approach trajectory and grasp pose are provided by an upstream policy or scripted motion.

During the pre-contact phase, the desired-force head predicts the signed normal-force target $\hat F^\star$. The gripper then closes incrementally while the wrench head continuously estimates the current normal force $\hat F_z(t)$ from RGB observations and robot state.

Under our sensor convention, compression corresponds to negative $F_z$. The grasp-termination rule is:
\begin{equation}
\begin{aligned}
\hat F_z(t) > \hat F^\star
&\Rightarrow \text{continue closing}, \\
\hat F_z(t) \leq \hat F^\star
&\Rightarrow \text{stop and hold}.
\end{aligned}
\end{equation}

Thus, the desired-force head predicts the grasp-level normal-force target, whereas the wrench head provides the instantaneous $F_z$ estimate used for closure control. Both are predicted without tactile input at deployment. Recorded tactile measurements are used only as offline references for wrench evaluation and visualization of the force-target crossing.

%%%%%%%%%%%%%%%%%%%%%%%%%%%%%%%%%%%%%%%%%%%%%%%%%%%%%%%%%%%%%%%%%%%%%%%%%%%%%%%%
\section{EXPERIMENT}
%%%%%%%%%%%%%%%%%%%%%%%%%%%%%%%%%%%%%%%%%%%%%%%%%%%%%%%%%%%%%%%%%%%%%%%%%%%%%%%%

\subsection{Dataset and Evaluation Protocol}

We evaluate ZeroTouch on real-robot grasp sequences containing wrist-camera RGB, gripper state, local gravity direction, and tactile measurements recorded with a DM-Tac W2 vision-based tactile sensor. The tactile data provide deformation and six-axis wrench supervision.

The dataset contains 270 grasp episodes and 15,316 synchronized samples: 187 episodes are used for training, 31 for validation, 46 for final testing, and six for diagnostic analysis.

Validation is used for model and checkpoint selection. It contains 12 held-out grasp configurations of training objects, 10 sponge-variation episodes, and 9 color/instance-variation episodes. The test set evaluates three distribution shifts: a seen-object/unseen-grasp condition using an off-center grasp of the pink bottle, an unseen physical object, and a contents/load shift.

All normalization statistics are computed from the training set only. The tactile wrench is used as training supervision and offline reference; forces and moments are reported in sensor-reported newtons and newton-meters, respectively.
\subsection{Offline Evaluation}

We quantitatively evaluate the three outputs of ZeroTouch: the instantaneous six-axis wrench, dense tactile deformation, and the grasp-level desired compression target. For each wrench component  $q \in \{F_x,F_y,F_z,M_x,M_y,M_z\}$, frame-wise absolute errors are first averaged within each episode and then across episodes, such that each grasp contributes equally to the reported MAE.

Dense tactile prediction is evaluated using normalized deformation MAE. Desired-force prediction is evaluated once per episode using absolute error and signed error $\hat F^\star-F^\star$. Under our negative-compression convention, a negative signed error indicates a more negative, stronger predicted compression target.

The architecture ablation is performed on the validation set, while the final model is reported on training, validation, and the held-out test set.

\subsection{Physical Grasping Protocol}

We evaluate three distribution shifts: an unseen can with different geometry and mass, an unseen off-center grasp on a familiar bottle, and a content/load shift obtained by filling a bottle with water. For all methods, the upstream policy provides the arm motion and target grasp configuration; methods differ only in gripper-closure control.

\section{RESULTS}

\subsection{Architecture Ablation}

The architecture selection uses the full 31-episode validation set. Table~\ref{tab:ablation} progressively introduces pooled visual features, spatial cross-attention, tactile-map supervision, and a dedicated wrench token.

Robot state alone obtains an $F_z$ MAE of $2.017$~N. Adding pooled RGB features reduces the error to $1.115$~N. Spatial attention further reduces the error to $0.828$~N, while adding tactile-map supervision reaches $0.752$~N. The complete model with a dedicated wrench token obtains the lowest validation error of $0.531$~N.

\begin{table}[t]
\centering
\caption{
Architecture ablation for normal-force prediction. Validation contains held-out objects and grasp configurations. $\Delta$ Val denotes error reduction relative to the state-only baseline.
}
\label{tab:ablation}
\setlength{\tabcolsep}{3.5pt}
\renewcommand{\arraystretch}{1.08}
\begin{tabular}{lccc}
\toprule
\textbf{Model} &
\textbf{Train} &
\textbf{Val} &
\textbf{$\Delta$ Val} \\
&
\multicolumn{2}{c}{\textbf{$F_z$ MAE (N)}} &
\textbf{(\%)} \\
\midrule
State only               & 0.651 & 2.017 & 0.0 \\
Pooled RGB + state        & 0.185 & 1.115 & 44.7 \\
Attention + FiLM          & 0.173 & 0.828 & 58.9 \\
+ tactile-map loss        & \textbf{0.108} & 0.752 & 62.7 \\
+ wrench token            & 0.117 & \textbf{0.531} & \textbf{73.7} \\
\bottomrule
\end{tabular}
\end{table}

\subsection{Quantitative Interaction Estimation}
Table~\ref{tab:full_eval} evaluates the final ZeroTouch model across all six wrench components, dense tactile deformation, and the grasp-level desired compression target.
%%%%%%%%%%%%%%%%%%%%%%%%%%%%%%%%%%%%%%%%%%%%%%%%%%%%%%%%%%%%%%%%%%%%%%%%%%%%%%%%
\begin{table}[t]
\centering
\caption{
Quantitative evaluation of the final ZeroTouch model. Errors are aggregated at the episode level. Validation is used for model selection, whereas Test is reserved for final generalization evaluation.
}
\label{tab:full_eval}
\setlength{\tabcolsep}{3.0pt}
\renewcommand{\arraystretch}{1.05}
\begin{tabular}{llccc}
\toprule
\textbf{Output} &
\textbf{Metric} &
\textbf{Train} &
\textbf{Val} &
\textbf{Test} \\
& &
\textbf{187} &
\textbf{31} &
\textbf{46} \\
\midrule

$F_x$ & MAE [N]
& 0.058 & 0.117 & 0.385 \\

$F_y$ & MAE [N]
& 0.059 & 0.150 & 0.292 \\

$F_z$ & MAE [N]
& 0.111 & \textbf{0.531} & \textbf{0.828} \\

$M_x$ & MAE [N\,m]
& 0.120 & 0.405 & 0.649 \\

$M_y$ & MAE [N\,m]
& 0.127 & 0.635 & 0.849 \\

$M_z$ & MAE [N\,m]
& 0.043 & 0.221 & 0.352 \\

\midrule

Deformation & MAE [normalized]
& 0.157 & 0.393 & 0.610 \\

\midrule

$F^\star$ & Episode MAE [N]
& 0.074 & 0.451 & 0.679 \\

$F^\star$ & Signed error [N]
& -0.015 & -0.348 & -0.490 \\

\bottomrule
\end{tabular}
\end{table}

Prediction errors increase from training to held-out evaluation across the wrench, deformation, and desired-force outputs. The degradation varies across wrench axes, motivating separate reporting of all six components rather than using $F_z$ as a proxy for overall wrench accuracy. The desired-force predictor exhibits a negative signed bias on validation and test data, indicating a tendency to predict stronger compression targets than the sensor-derived references.

\subsection{Six-Axis Contact Estimation}

\begin{figure}[t]
    \centering
    \includegraphics[width=\linewidth]{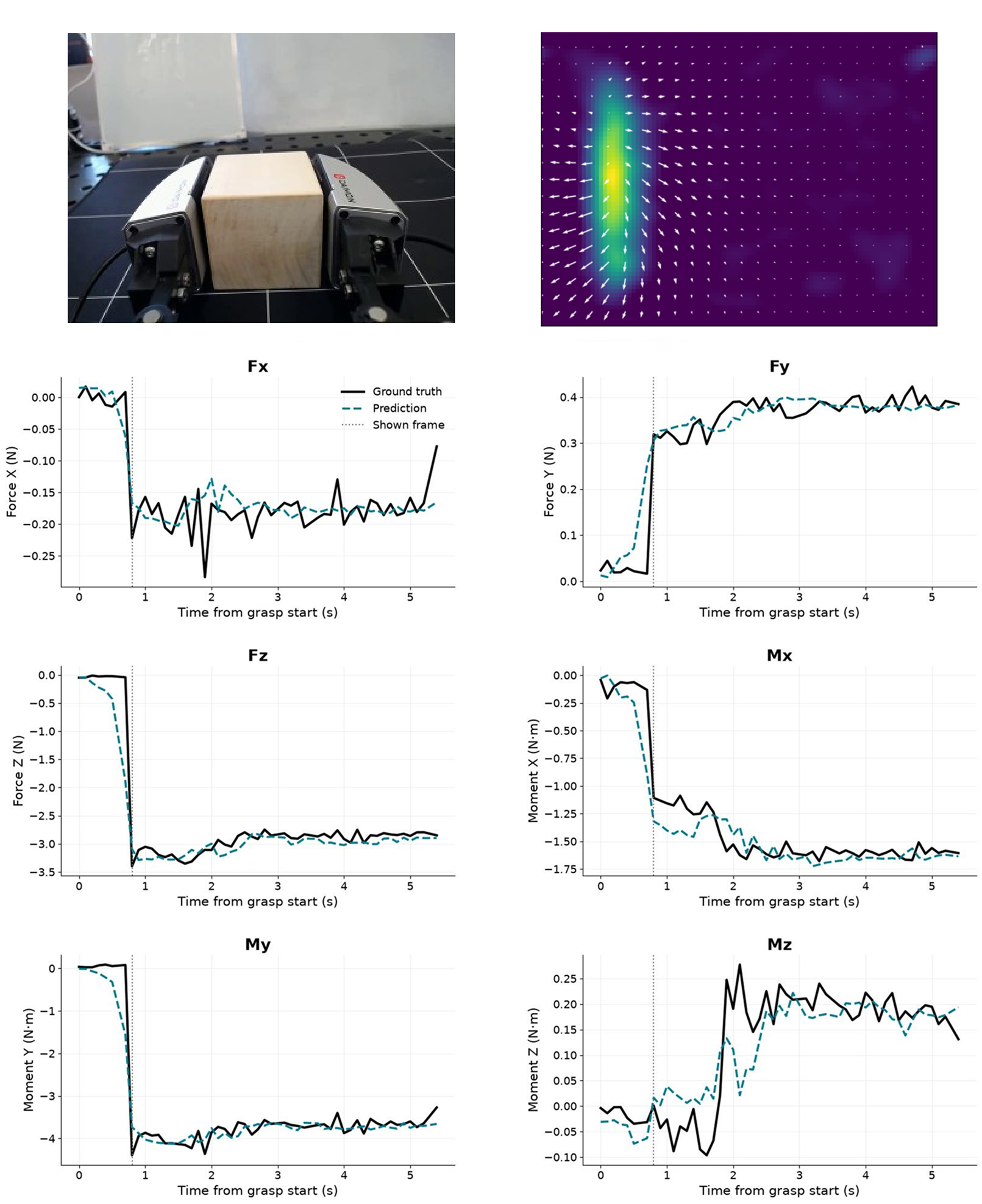}
    \caption{
    Six-axis contact estimation for a representative grasp.
    The example shows the wrist RGB observation, tactile deformation,
    and recorded and predicted force and moment components.
    The tactile sensor is used as an offline reference and is not an
    inference input to ZeroTouch.
    }
    \label{fig:six_axis}
    \vspace{-3mm}
\end{figure}

Although grasp termination uses $F_z$, ZeroTouch predicts the complete six-axis wrench. Fig.~\ref{fig:six_axis} shows that the predicted forces and moments follow the recorded interaction, while the dense tactile output provides complementary spatial information about the contact distribution.

\subsection{Visual Desired-Force Prediction and Grasp Termination}

\begin{figure}[t]
    \centering
    \includegraphics[width=0.8\linewidth]{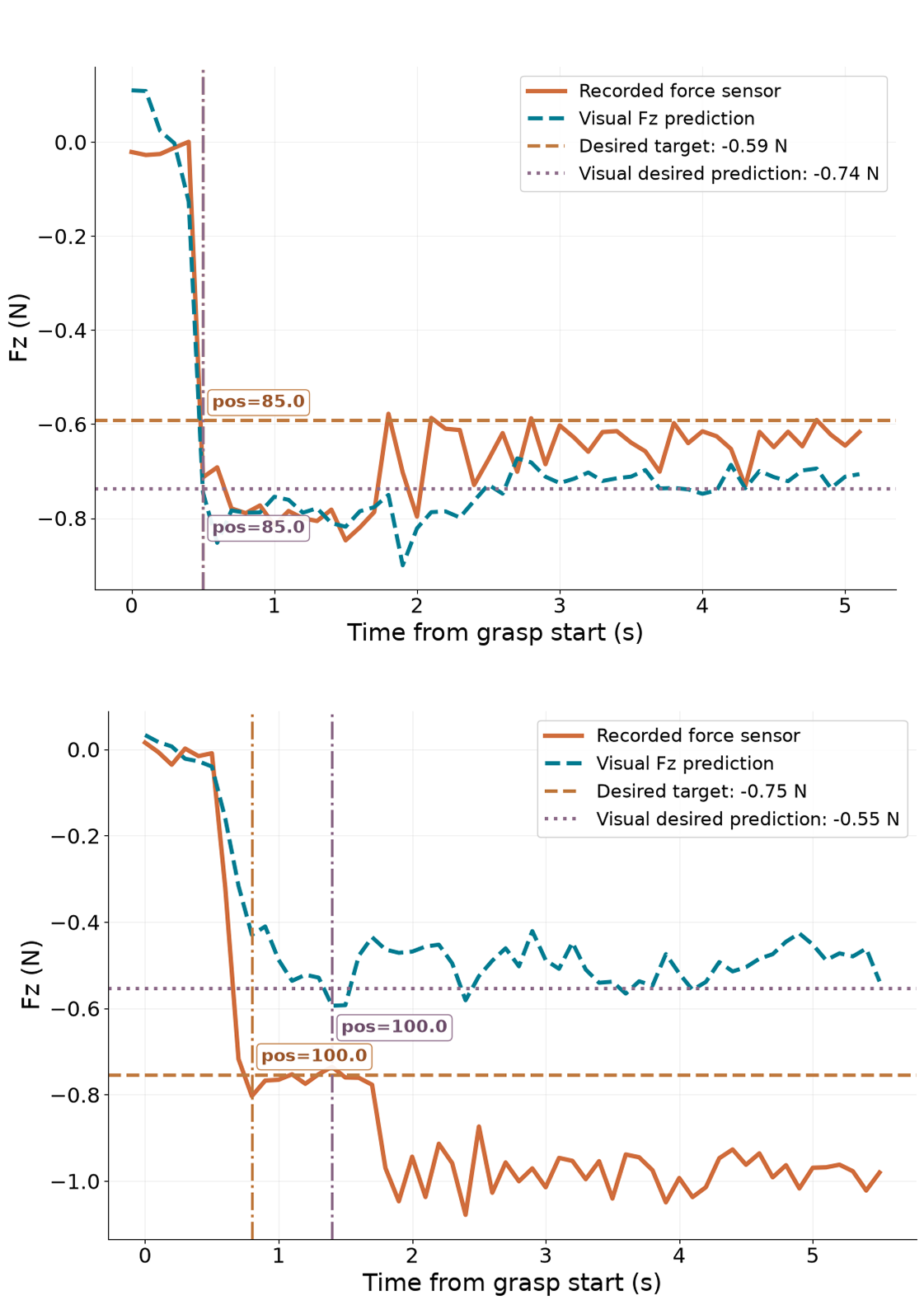}
    \caption{
    Current-force and desired-force prediction for two representative grasp
    episodes. Solid curves show the recorded normal force, dashed curves show
    the visual force prediction, and horizontal lines denote the recorded
    and predicted desired compression targets. Vertical markers indicate the
    corresponding threshold crossings. The examples illustrate the relative
    force-target criterion used by ZeroTouch; closed-loop grasp termination
    is evaluated separately in the physical robot experiments.
    }
    % \vspace{-4mm}
    \label{fig:force_crossing}
\end{figure}

Figure~\ref{fig:force_crossing} illustrates how the two visual predictions are combined by the proposed termination rule. ZeroTouch estimates both the instantaneous compression $\hat F_z(t)$ and the grasp-level desired compression $\hat F^\star$.

In both examples, the individual force and desired-force predictions exhibit visible offsets from their sensor references. Nevertheless, the predicted force reaches the predicted target near the corresponding sensor-reference crossing. Thus, errors in $F_z$ and $F^\star$ do not necessarily translate directly into the same error in the relative force-target criterion used by the controller.

We use these trajectories only to illustrate the termination mechanism. Quantitative prediction accuracy is reported separately for $F_z$ and $F^\star$ in Table~\ref{tab:full_eval}, while grasp termination is evaluated in the physical robot experiments.

\begin{figure*}[t]
    \centering
    \includegraphics[width=\textwidth]{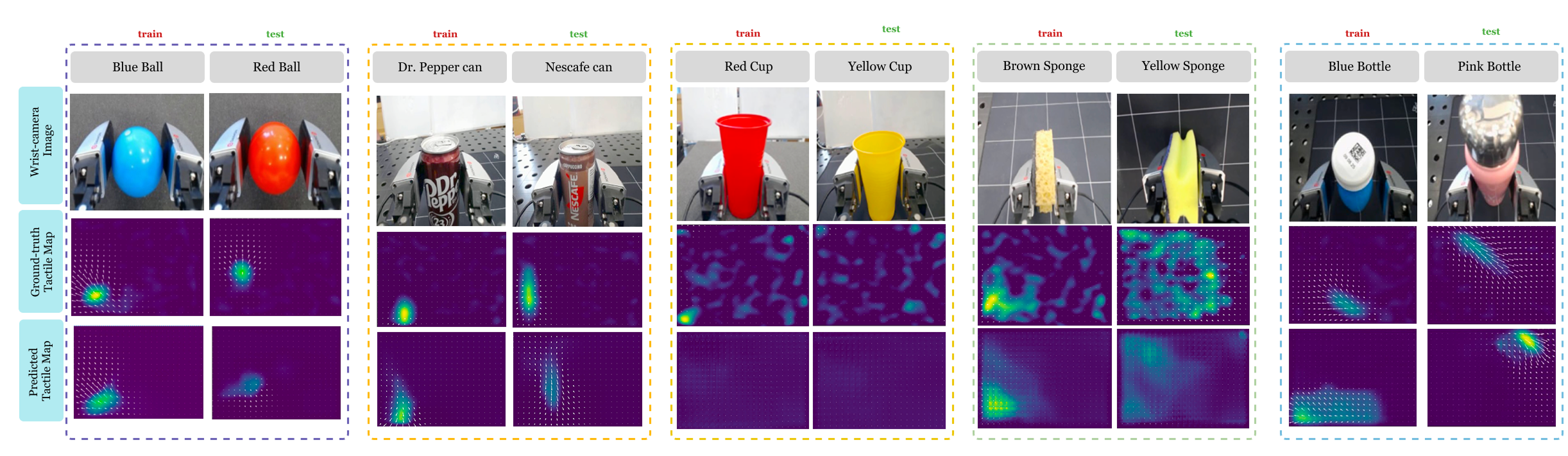}
    \caption{
    Generalization of tactile deformation prediction to unseen objects and
    held-out grasp configurations. Rows show RGB observations, recorded
    tactile deformation, and predicted deformation. The examples include
    both objects excluded from training and objects seen during training
    but evaluated in previously unseen poses and grasp configurations.
    }
    % \vspace{-4mm}
    \label{fig:train-test}
\end{figure*}

\subsection{Generalization to Unseen Objects and Grasp Configurations}

We next examine whether the learned spatial contact representation transfers beyond the exact objects used for training.

Figure~\ref{fig:train-test} shows representative training and held-out interactions. Rows show the wrist RGB observation, recorded tactile deformation, and predicted deformation. Held-out examples include both objects excluded from training and training objects evaluated under unseen poses or grasp configurations. ZeroTouch preserves the approximate contact location and dominant deformation structure across these variations. These examples complement Table~\ref{tab:full_eval}, whose validation errors are computed over all 31 episodes rather than selected frames.

\begin{figure}[t]
    \centering
    \includegraphics[width=1\linewidth]{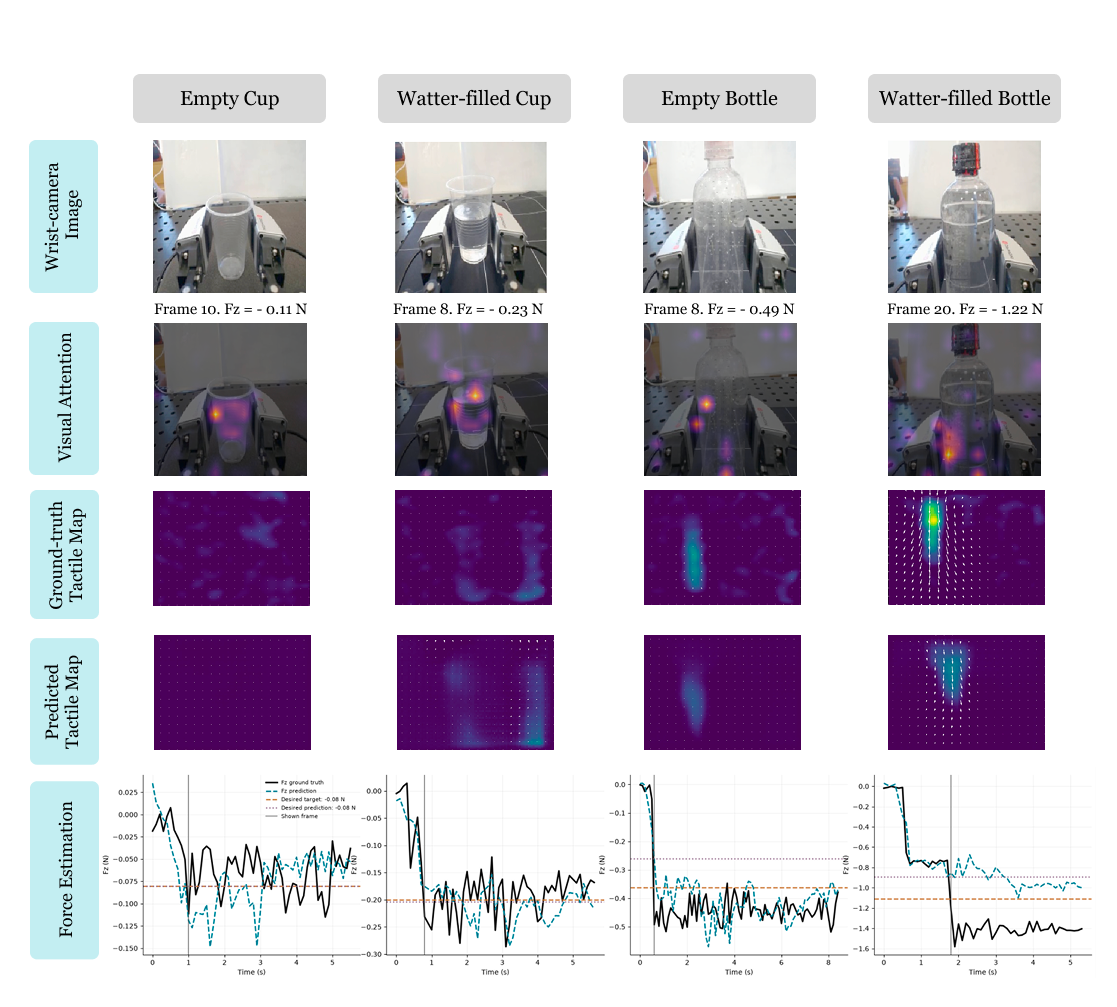}
    \caption{
    Effect of object loading on predicted contact.
    Columns compare empty and loaded cups and bottles.
    Rows show the RGB input, visual attention, recorded tactile
    deformation, predicted deformation, and normal-force response.
    }
    \vspace{-5mm}
    \label{fig:loading}
\end{figure}

\subsection{Sensitivity to Object Load and Grasp Configuration}

The observed interaction force can vary not only across objects but also across loading conditions and grasp configurations. We qualitatively examine these two sources of variation.

Figure~\ref{fig:loading} shows representative interactions with empty and water-filled containers. The recorded normal-force trajectories differ between the paired examples. Because these examples may also differ in grasp placement and closure execution, they should not be interpreted as an isolated causal measurement of the effect of object load.

In the cup examples, the model predicts a larger normal-force magnitude for the water-filled condition than for the empty condition. The visualized cross-attention weights also differ between the two interactions; however, these weights are presented only as a qualitative inspection of the learned representation and do not establish which visual cues determine the prediction.

The bottle examples illustrate a limitation of the proposed formulation. Internal load may be weakly observable, or unobservable, from the pre-contact RGB image and proprioceptive state. Consequently, the predicted grasp-compression target may differ from the target derived from the tactile-sensor reference. This example indicates that tactile-free prediction cannot reliably recover physical properties that produce insufficient observable evidence.

\begin{figure*}[t]
    \centering
    \includegraphics[width=0.8\linewidth]{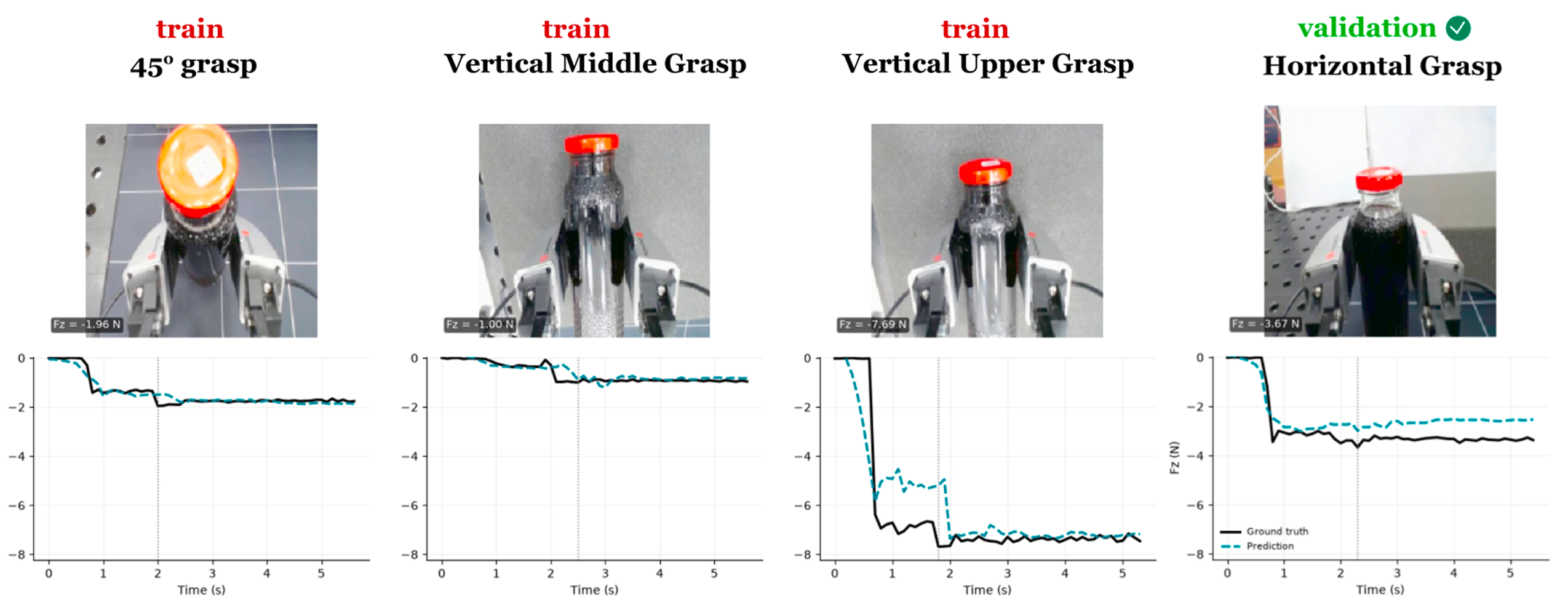}
    \caption{
    Normal-force prediction for the same bottle across four grasp
    configurations. Changing grasp location and orientation changes the
    recorded interaction even though object identity is fixed.
    }
    \label{fig:pose_dependence}
\end{figure*}

Figure~\ref{fig:pose_dependence} isolates a different factor by keeping the object fixed while changing grasp orientation and contact location. The same bottle is grasped horizontally, at an angle, and at different vertical locations.

The resulting normal-force response changes markedly between configurations. In particular, grasping an object away from its central load-bearing region introduces a larger lever arm between the contact and the supported mass. Consequently, the fingers must resist both translation and a stronger rotational tendency.

ZeroTouch follows these configuration-dependent changes despite the object identity remaining unchanged. This result motivates the use of spatial visual features together with gripper state and local gravity direction rather than assigning a single characteristic force to each object.

\subsection{Representation Analysis}

\begin{figure}[t]
    \centering
    \includegraphics[width=\linewidth]{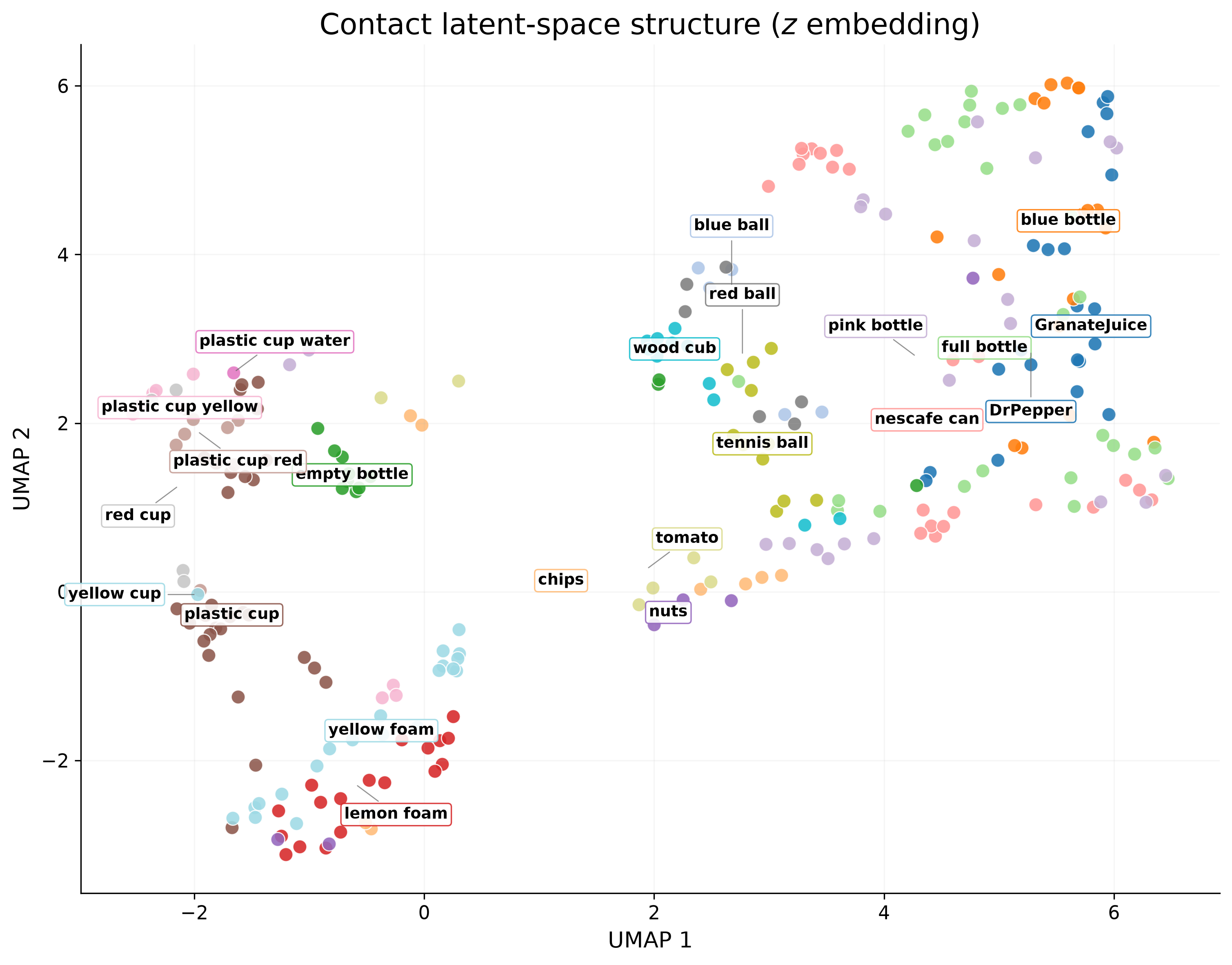}
    \caption{
    UMAP visualization of the learned interaction latent space.
    Each point represents the median contact latent of one grasp episode.
    Labels and train/validation membership are used only for post-hoc
    visualization and are not inputs to UMAP.
    }
    \label{fig:latent}
\end{figure}

To inspect the representation learned by the contact decoder, we extract the latent vector $\mathbf{z}$ immediately before the wrench head. Latents are collected from contact frames and aggregated within each episode using their coordinate-wise median. Thus, each point in Fig.~\ref{fig:latent} represents one complete grasp episode rather than an individual video frame.

We project the episode representations to two dimensions using UMAP with cosine distance. Object identities, semantic groups, and train/validation membership are not provided to UMAP and are used only for visualization after the projection.

The resulting embedding exhibits visible organization across several interaction types. In particular, groups of compliant objects, rigid containers, and related grasp configurations occupy different regions of the projection. We use this visualization only as a qualitative analysis of the learned representation; the quantitative generalization results are reported independently in the preceding experiments.

\subsection{Physical Grasping Experiments}

Finally, we evaluate grasp execution under three distribution shifts: an unseen object, a seen-object/unseen-grasp configuration, and a contents/load shift. Each method is evaluated over 20 trials per condition. A trial is counted as successful if the object is lifted by $5$~cm, held remains securely held for $2$~s without being dropped.

The unseen-object experiment uses a Nescafe can that is absent from training and validation. The seen-object/unseen-grasp experiment uses the pink bottle, which is represented during training through grasps near its central region, while evaluation uses a held-out off-center grasp close to its end. The contents/load experiment uses a water-filled bottle, while training contains the corresponding empty-container condition.

OpenVLA and SmolVLA directly predict gripper actions as part of their manipulation policies, whereas ZeroTouch regulates gripper closure using the visually estimated current compression and a grasp-dependent desired-force target.

ZeroTouch achieves $19/20$ successful grasps ($95\%$) in the unseen-object condition, $16/20$ ($80\%$) in the seen-object/unseen-grasp condition, and $18/20$ ($90\%$ under the content/load shift). OpenVLA achieves $5/20$ ($25\%$), $8/20$ ($40\%$), and $11/20$ ($55\%$), while SmolVLA achieves $2/20$ ($10\%$), $5/20$ ($25\%$), and $7/20$ ($35\%$), respectively.

Under the evaluated conditions, ZeroTouch achieves higher grasp success than both VLA baselines across all three distribution shifts.

\begin{table}[t]
\centering
\caption{
Physical grasping success over 20 trials per condition.
}
\label{tab:physical_grasping}
\setlength{\tabcolsep}{3.0pt}
\renewcommand{\arraystretch}{1.08}
\begin{tabular}{lccc}
\toprule
\textbf{Method} &
\textbf{Unseen object} &
\textbf{Unseen grasp} &
\textbf{Content shift} \\
\midrule
SmolVLA
& 2/20 (10\%) & 5/20 (25\%) & 7/20 (35\%) \\

OpenVLA
& 5/20 (25\%) & 8/20 (40\%) & 11/20 (55\%) \\

ZeroTouch 
& \textbf{19/20 (95\%)} &
  \textbf{16/20 (80\%)} &
  \textbf{18/20 (90\%)} \\
\bottomrule
\end{tabular}
\vspace{-4mm}
\end{table}

\section{LIMITATIONS}

The experiments reveal several limitations of the current formulation. First, physical properties that produce little observable evidence remain difficult to infer before contact. This is particularly apparent under the content/load shift, where similar external appearances can correspond to different sensor-derived compression targets.

Second, $F^\star$ is an empirical desired compression obtained from the settled force of a grasp that satisfies the lift-and-hold protocol. It should not be interpreted as a globally optimal or theoretically minimum-safe grasp force.

Third, prediction accuracy is not uniform across the six wrench components. Several moment components degrade more strongly than $F_z$ under held-out conditions.

Fourth, because grasp termination depends on the relative crossing of the predicted current compression and desired-compression target, errors in $F_z$ and $F^\star$ may partially cancel in individual trajectories. A successful threshold crossing should therefore not be interpreted as evidence that either quantity is individually accurate. The two outputs are evaluated separately in addition to their combined termination criterion.

Finally, the tactile sensing system provides the wrench reference used for training and offline evaluation; we do not provide an independent external force/torque calibration. The held-out test set also covers a limited number of objects, grasp configurations, and load shifts. Broader evaluation is required to characterize the operating range of the method.

\section{CONCLUSION}

We introduced ZeroTouch, a tactile-supervised vision-and-proprioception model that predicts a dense contact representation, an instantaneous six-axis interaction wrench, and a grasp-level desired compression without requiring tactile input at deployment.

Validation is used for model and checkpoint selection, whereas the test set is reserved for three distribution shifts: a seen-object/unseen-grasp condition, an unseen physical object, and a content/load shift. The architecture ablation reduces validation $F_z$ MAE from $2.017$~N for the state-only baseline to $0.531$~N. The final evaluation reports all six wrench components, spatial contact prediction, and desired-force estimation.

The current-force and desired-force predictions are used jointly for grasp termination: closure stops when the visually estimated compression reaches the predicted grasp-dependent target. Representative trajectories show that errors in the individual estimates do not necessarily produce an equivalent displacement of their relative crossing. Closed-loop performance is therefore evaluated directly in the physical grasping experiments.

Across 20 trials per condition, ZeroTouch achieves $95\%$ grasp success in the unseen-object condition, $80\%$ in the seen-object/unseen-grasp condition, and $90\%$ under the content/load shift. Under the same evaluation protocol, OpenVLA achieves $25\%$, $40\%$, and $55\%$, while SmolVLA achieves $10\%$, $25\%$, and $35\%$, respectively. Under the evaluated conditions, these results indicate that tactile supervision can improve a vision-and-proprioception grasp controller without requiring tactile input at deployment. Remaining limitations include visually ambiguous object loading, reduced accuracy for some wrench components, and the limited scope of the current held-out evaluation.

\end{document}